\PassOptionsToPackage{dvipsnames}{xcolor}
\documentclass{template/bytedance}

\usepackage[utf8]{inputenc} 
\usepackage{amsfonts}       
\usepackage{nicefrac}       

\usepackage{amsmath} 
\usepackage{enumerate}
\usepackage{enumitem}
\usepackage{wrapfig}
\usepackage{graphicx}
\usepackage{rotating}
\usepackage{colortbl}
\usepackage{float}
\usepackage{placeins}
\usepackage{wrapfig}
\usepackage{pgfplots}
\usepackage{subcaption}
\usepackage{graphicx}
\pgfplotsset{compat=1.16}
\usetikzlibrary{arrows.meta,positioning,shapes.geometric,calc,fadings,decorations.pathmorphing}

\newcommand{\TheName}{\textsc{SEAL}}

\fancypagestyle{firststyle}{%
  \fancyhf{}%
  \fancyhead[L]{\vskip 4mm\includegraphics[height=8mm]{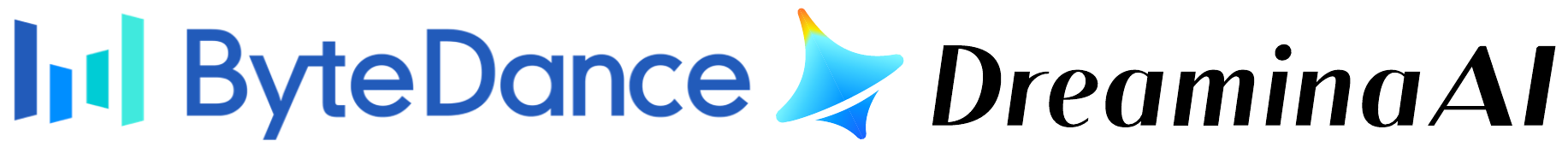}}%
}

\definecolor{autoskillblue}{RGB}{220,232,250}   
\definecolor{autoskilldeep}{RGB}{44,90,181}     
\definecolor{autoskilldark}{RGB}{24,55,130}     

\definecolor{codexamber}{RGB}{224,139,60}
\definecolor{codexamberlight}{RGB}{255,221,184}
\definecolor{codexamberdark}{RGB}{162,93,28}
\definecolor{bdteal}{RGB}{69,191,199}
\definecolor{bdteallight}{RGB}{198,233,235}
\definecolor{bdtealdark}{RGB}{36,128,134}
\definecolor{deltagreen}{RGB}{22,128,57}

\usepackage{pifont}
\usepackage{wasysym}

\definecolor{markgreen}{RGB}{34,139,52}
\definecolor{markred}{RGB}{200,40,40}
\definecolor{markamber}{RGB}{210,150,0}

\lstdefinestyle{jsonstyle}{
    basicstyle=\ttfamily\footnotesize,
    breaklines=true,
    breakatwhitespace=false,
    columns=fullflexible,
    keepspaces=true,
    frame=single,
    framesep=4pt,
    xleftmargin=0pt,
    xrightmargin=0pt,
    linewidth=\linewidth,
    postbreak=\mbox{\textcolor{gray}{$\hookrightarrow$}\space}
}

\title{\TheName: Can Saturated Benchmarks Be Revived by LLM-as-a-Meta-Judge?}

\author[1,2,*]{Jiamin Chen}
\author[1,*]{Yidi Wu}
\author[1]{Qiexiang Wang}
\author[1]{Qianben Chen}
\author[2]{Yuchen Li}
\author[2]{Yansen Zhang}
\author[2]{Xiaokun Zhang}
\author[1,\dagger]{Wangchunshu Zhou}
\author[2]{Chen Ma}

\affiliation[1]{ByteDance Inc.}

\affiliation[2]{City University of Hong Kong}

\contribution[*]{Equal Contribution}
\contribution[\dagger]{Corresponding author}

\abstract{
Widely used language-model benchmarks are increasingly saturated, with frontier systems often receiving near-tied scores that standard metrics cannot resolve. Rather than constructing harder alternatives, we ask whether existing tasks can be made informative again through improved evaluation over the same candidate outputs.
Therefore, we present \emph{S}eeded \emph{E}limination with \emph{A}daptive \emph{L}LM-as-a-Meta-Judge, a self-improving evaluation protocol for extracting latent ranking signal from saturated benchmarks.
\TheName{} seeds candidate outputs into a single elimination and evaluates each match with task-level principles plus self-improving checklist criteria.
We evaluate \TheName{} on multiple saturated benchmarks covering code generation, mathematical reasoning, knowledge-intensive question answering, and tool-use agent task completion.
Across these settings, \TheName{} improves the ranking-accuracy--latency trade-off over competing protocols, attaining 0.83--1.00 Spearman agreement with full pairwise judging and 4/4 top-1 agreement, while requiring only 11.89 calls per task compared with 28.00 for full pairwise evaluation.
}

\date{\today}
\github{\url{https://github.com/jiaminchen-1031/SEAL}}
\correspondence{Wangchunshu Zhou, \email{chunshu@bytedance.com}}

\begin{document}

\maketitle

\section{Introduction}
Benchmarks~\citep{wang-etal-2018-glue,liang2023helm} have been central to measuring progress in large language models (LLMs), but many once-discriminative benchmarks now operate near their ceiling for frontier systems~\citep{wang2019superglue,kiela-etal-2021-dynabench,akhtar2026benchmarksplateau}.
A natural response is to develop harder benchmarks, which remains essential for measuring emerging capabilities but requires additional task design, validation, maintenance, and repeated evaluation.
We ask a complementary question: given a fixed set of benchmark tasks and already-generated candidate outputs, are saturated benchmarks truly exhausted, or can they be made informative again by changing only the evaluation protocol?
Our key observation is that saturation is not only a property of the task distribution, but also a consequence of limited evaluation resolution. 
Top systems may receive the same task-level outcome, yet their outputs can still differ in latent quality. 
Coarse metrics and generic pointwise judges often fail to expose these differences, especially when they are affected by evaluator or response-length biases~\citep{wang2023llmfair,dubois2024lengthcontrolledalpacaeval}.
Pairwise preference evaluation can expose such latent ranking signal in open-ended LLM evaluation, but the challenge is to recover it without paying the latency of exhaustive all-pairs comparison~\citep{chiang2024chatbotarena}.

Motivated by this view, we introduce \TheName{}, \emph{S}eeded \emph{E}limination with \emph{A}daptive \emph{L}LM-as-a-Meta-Judge.
\TheName{} first uses a cheap listwise judge to seed candidate outputs, then ranks them through a single-elimination tournament.
Each match is judged pairwise under predefined task-level principles, while an LLM meta-judge generates rank-adaptive checklist items that become finer in later rounds.
In this way, \TheName{} keeps aggregation semantics stable while increasing evaluation resolution only for close comparisons among strong candidates.
Compared with the regime where one-shot judging is too noisy but exhaustive all-pairs judging is too slow, \TheName{} aims for a better ranking-accuracy--latency compromise: recovering much of the signal of exhaustive pairwise evaluation while retaining the lower-latency profile of tournament-style protocols.

\textbf{Our contributions are as follows:}

\begin{itemize}[leftmargin=*]
    \item We formulate saturated-benchmark evaluation as a fixed-candidate re-ranking problem and propose \TheName{}, a seeded-elimination protocol for making existing benchmarks informative again.
    \item We introduce an LLM-as-a-Meta-Judge mechanism that adaptively refines checklist criteria across tournament rounds while fixed task-level principles preserve stable aggregation.
    \item Across HumanEval, GSM8K, MMLU, and BFCL-v2, we show that \TheName{} provides a better ranking-accuracy--latency compromise than pointwise, listwise, and fixed-bracket baselines. It approaches exhaustive pairwise agreement with substantially lower evaluation overhead in both cost and time.
    
\end{itemize}


\begin{figure*}[t]
  \centering
  \includegraphics[width=1\textwidth]{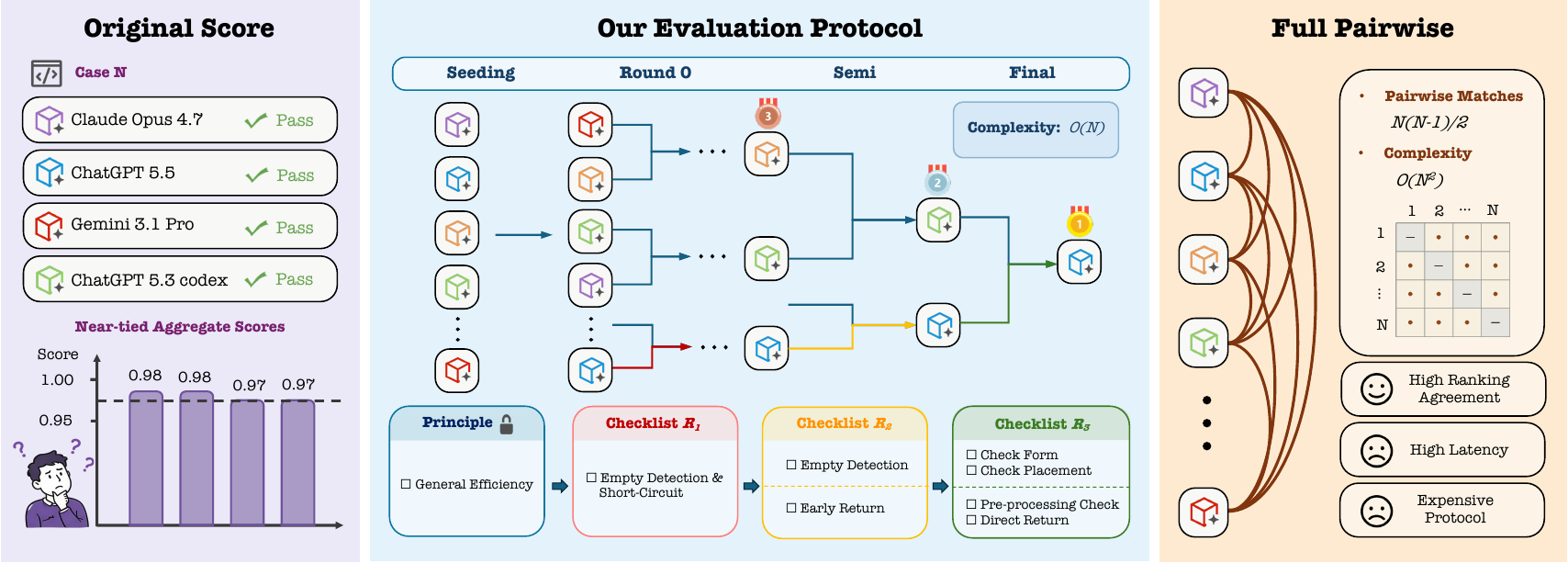}
  \vspace{-1mm}
  \caption{
  \TheName{} restores ranking resolution through seeded elimination, fixed principles, and progressively evolved checklists, achieving interpretable re-ranking with much lower cost than full pairwise comparison.
  }
  \label{fig:method_overview}
  \vspace{-4mm}
\end{figure*}

\section{\TheName{}: Saturated Re-Ranking with LLM-as-a-Meta-Judge}\label{sec:method}
Let $x$ denote a benchmark task and $\mathcal{Y}=\{y_1,\ldots,y_N\}$ the candidate outputs from a fixed set of systems.
A native benchmark metric maps each output to a coarse task-level outcome.
We call a task \emph{saturated} when this metric leaves insufficient resolution within the candidate pool, e.g., when many systems receive the same outcome or when top-tier aggregate scores fall within a narrow band~\citep{akhtar2026benchmarksplateau}.
Our goal is not to replace the native metric, but to recover additional ranking signal in these saturated regions.
\TheName{} produces task-level rankings and aggregates them into a benchmark-level ranking, while targeting a better ranking-accuracy--latency trade-off than one-shot or exhaustive pairwise judging..
 
\subsection{Seeded Elimination}

\TheName{} is a seeded-elimination protocol with an LLM-as-a-Meta-Judge.
For each saturated task, it first applies a cheap listwise judge to assign candidate outputs into coarse seed tiers.
These seeds initialize a single-elimination bracket, spreading likely strong candidates across the tournament and reducing early elimination among close competitors.
This design follows the broader idea that rankings can be recovered from informative subsets of pairwise comparisons rather than from enumerating all possible pairs~\citep{jamieson2011activeranking,wauthier2013efficientranking}.
Each match is judged pairwise using a two-layer rubric: fixed task-level principles that remain constant across rounds, and rank-adaptive checklist items produced by the LLM meta-judge.
Early rounds use broad checklist items to separate obvious differences, while later rounds use finer checklist items for close comparisons among strong candidates.

\subsection{LLM-as-a-Meta-Judge}

The meta-judge does not directly assign final scores or rankings.
Instead, it improves the evaluation procedure by refining the checklist used in subsequent matches.
This differs from meta-judging for self-improving alignment, where the meta-judge is used to improve reward or preference learning signals~\citep{wu2025metarewarding}.
After each tournament round, the protocol summarizes match evidence and asks the meta-judge to identify which fixed principles remain under-resolved.
It then proposes finer checklist items grounded in those principles and tailored to the remaining candidate pool.
This implements a rank-resolution principle: different regions of the ranking require different rubric granularity.
By adapting only the checklist layer while keeping the principle layer fixed, \TheName{} improves resolution for difficult comparisons without changing the semantics of benchmark-level aggregation.


\begin{table*}[t]
\centering
\small
\setlength{\tabcolsep}{4pt}
\begin{tabular}{lcccccccc}
\toprule
& \multicolumn{4}{c}{Agreement with FullPair ($\rho$)}
& \multicolumn{4}{c}{Resolution Gain over Original ($\times$)} \\
\cmidrule(lr){2-5} \cmidrule(lr){6-9}
Protocol & HumanEval & GSM8K & MMLU & BFCL-v2
& HumanEval & GSM8K & MMLU & BFCL-v2 \\
\midrule
Original  & 0.43 & -0.57 & 0.79 & 0.71 & 1.00 & 1.00 & 1.00 & 1.00 \\
Pointwise & \textbf{0.95} & 0.81 & 0.86 & 0.93 & 1.18 & 5.29 & 1.13 & 0.70 \\
FixedRub  & 0.10 & 0.26 & \textbf{0.95} & 0.88 &  0.12 & 5.72 & 0.77 & 0.79 \\
Listwise  & 0.69 & 0.81 & 0.71 & 0.95 & 9.84 & 28.64 & 3.52 & 1.83 \\
FlatBrk   & 0.93 & 0.57 & 0.93 & 0.83  & 1.77 & 9.87 & 1.30 & 0.54 \\
\rowcolor{gray!40}
\TheName{} & \textbf{0.95} & \textbf{0.83} & \textbf{1.00} & \textbf{0.98} & 2.10 & 10.50 & 1.30 & 0.80 \\
FullPair  & 1.00 & 1.00 & 1.00 & 1.00 & 1.88 & 30.07 & 1.77 & 0.64 \\
\bottomrule
\end{tabular}
\vspace{-3pt}
\caption{
Main ranking results.
Ranking agreement with full pairwise judging, measured by Spearman's rank correlation $\rho$.
Resolution gain is the average score separation across all candidates divided by the native metric.
}
\vspace{-4mm}
\label{tab:main-results}
\end{table*}
 

\subsection{Aggregation and Latency}

For each pairwise match, the judge returns structured votes over the fixed principles and current checklist items. The backend computes the match winner from these votes rather than relying on an unconstrained overall score.
After all matches for a task finish, \TheName{} converts the bracket into a full task-level ranking. 
Candidates eliminated in the same round are ordered by their accumulated principle-level vote margin across the matches they played.
Benchmark-level rankings aggregate the resulting task-level ranks across saturated tasks using a normalized Borda score. Remaining ties are broken by the mean principle-level margin.
For $N$ candidates on a task, exhaustive pairwise judging requires $\binom{N}{2}$ comparisons, while \TheName{} uses one seeding call and $N-1$ tournament matches, plus a small number of meta-judge refinement calls.
Thus, \TheName{} is designed to occupy the useful middle ground between noisy one-shot judging and expensive all-pairs comparison.
 
\section{Experiments}
\label{sec:experiments}
We evaluate whether \TheName{} provides a better ranking-accuracy--latency trade-off on saturated benchmark regions.
Our experiments address three questions:
(1) whether \TheName{} agrees with a high-resolution pairwise reference;
(2) whether it reduces judging latency relative to exhaustive pairwise comparison; and
(3) whether adaptive meta-judge refinement improves over a non-adaptive tournament backbone.

\noindent\textbf{Benchmarks and Models.}
We evaluate on four benchmarks spanning different task settings: HumanEval~\citep{chen2021evaluatingcode} for code generation, GSM8K~\citep{cobbe2021trainingverifiers} for mathematical reasoning, MMLU~\citep{hendrycks2021mmlu} for knowledge-intensive multiple-choice question answering, and BFCL-v2~\citep{patil2025bfcl} for function-calling/tool-use evaluation.
For each benchmark, we start from a public leaderboard and select 8 strong candidate models whose original leaderboard scores are already close enough that the native metric provides limited separation, with details described in Section~\ref{app:leaderboards-models}.

\noindent\textbf{Baselines.}
We compare \TheName{} against six automatic evaluation protocols.
\textsc{Original} ranks systems by the native benchmark metric.
\textsc{Pointwise} independently scores each candidate output with an LLM judge.
\textsc{FixedRub} adds a fixed rubric before pointwise scoring.
\textsc{Listwise} asks an LLM judge to rank all candidate outputs in one call.
\textsc{FlatBrk} uses the same tournament backbone as \TheName{} but without adaptive checklist refinement.
\textsc{FullPair} exhaustively compares all candidate pairs and serves as the high-resolution reference when available.
They cover common LLM-as-a-judge settings, including pointwise scoring, rubric-guided evaluation, listwise ranking, and pairwise preference comparison~\citep{liu2023geval,zheng2023llmjudge,kim2024prometheus}.

The detailed experimental settings including LLM choices, predefined principles, checklists, and prompt templates are provided in Appendix~\ref{app:prompts}.

\subsection{Agreement with Pairwise Judging}

Table~\ref{tab:main-results} reports ranking agreement with full pairwise judging.
\TheName{} achieves consistently high agreement with \textsc{FullPair}: $0.95$ on HumanEval, $0.83$ on GSM8K, $1.00$ on MMLU, and $0.98$ on BFCL-v2.
This places \TheName{} at the best non-exhaustive protocol on every benchmark.
In terms of resolution gain, \textsc{SEAL} increases the discriminative spread over the original metric on most benchmarks. However, it reflects the discriminative spread of candidate scores, rather than the validity of the induced ranking.
%
We additionally report top-1 agreement with \textsc{FullPair}, together with the exact leaderboard snapshot for the candidate pool in Section~\ref{app:leaderboards-models}.
As illustrated, \TheName{} achieves $4/4$ top-1 agreement.
This result indicates that the tournament approximation preserves the most consequential leaderboard decision even though it does not evaluate every candidate pair.
The stability analysis in Section~\ref{app:stability} further confirms that \TheName{} produces stable rankings under both task subsampling and repeated full runs.

\subsection{Accuracy--Latency Trade-off}
\begin{wrapfigure}{r}{0.5\textwidth}
    \centering
    \includegraphics[width=\linewidth]{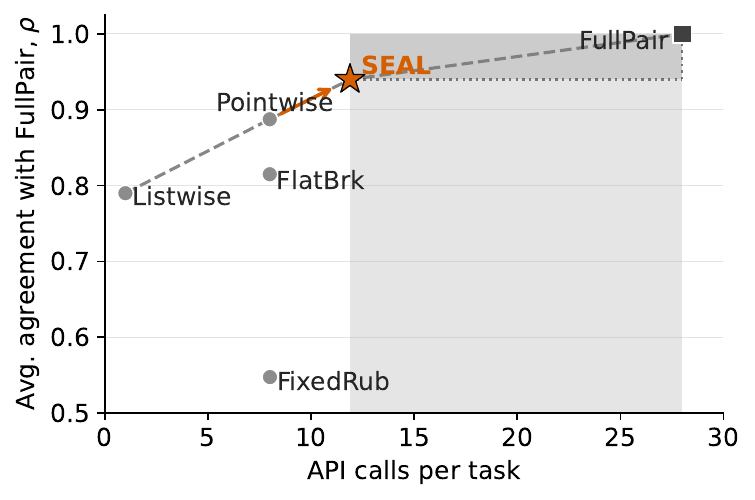}
    \vspace{-7mm}
    \caption{Accuracy--latency trade-off of automatic judging protocols.
    \TheName{} lies near the high-agreement, low-latency region by combining a tournament backbone with adaptive checklist refinement.}
    \vspace{-4mm}
    \label{fig:accuracy-latency}
\end{wrapfigure}

Figure~\ref{fig:accuracy-latency} visualizes the trade-off between agreement with \textsc{FullPair} and the number of sequential calls per task for 8 candidates.
The one-shot protocols are cheaper, but their ranking quality is less stable across
benchmarks.
\textsc{Listwise} uses only a single call per task and can produce large resolution gains, but its
agreement drops on HumanEval and MMLU.
\textsc{Pointwise} and \textsc{FixedRub} scale linearly in the number of candidates, but their absolute scores remain vulnerable to score compression and rubric-insensitive ties in saturated regions.

\TheName{}
uses a near-linear tournament backbone rather than the quadratic number of comparisons required by \textsc{FullPair}, while spending a small number of additional calls on checklist evolution when the remaining candidates become difficult to separate.
In general, \TheName{} uses $11.89$ calls per task compared with $28.00$ calls for \textsc{FullPair}, a reduction of about $57.5\%$, while reaching the same Spearman agreement as the strongest non-exhaustive baseline.
Thus, the adaptive tournament does
provide a substantially better agreement--latency compromise 
than other baselines.
%
The detailed cost profile including token consumption and actual cost can be found in Section~\ref{app:dollar-cost}.

\subsection{Case Study: Adaptive Refinement vs.\ Non-Adaptive Backbone}
Since \textsc{Flat-Bracket} shares \TheName{}'s bracket structure and only disables checklist evolution, its gap from \TheName{} isolates the effect of the adaptive meta-judge.
On HumanEval/92 (\texttt{any\_int}), the seed checklist contains only a generic predicate, ``handles spec details not directly tested,'' under $P_1$.
\TheName{} refines this into concrete checks for observed failure modes: whether validation uses \texttt{isinstance}, checks all three arguments, rejects whole-valued floats such as \texttt{3.0}, and avoids silent coercion with \texttt{int($\cdot$)}.
These finer checklist items yield decisive per-item verdicts, whereas the original seed predicate leaves the same candidates coarsely indistinguishable.

\subsection{Discussions}
Beyond the main accuracy--latency results, we examine whether \TheName{} is reliable enough to serve as a practical saturated-benchmark re-ranking protocol.
The analysis below focuses on three complementary concerns: 1) whether the resulting leaderboards remain stable under task subsampling and repeated judge runs, 2) how the selected saturated candidate pools affect leaderboard conclusions, and 3) what evaluation budget is required in terms of API calls, token consumption, and estimated dollar cost.
Together, these results show that \TheName{} not only approaches exhaustive pairwise judging in ranking agreement, but also produces reproducible leaderboard decisions with substantially lower operational cost.

\subsubsection{Stability Analysis}
\label{app:stability}
We additionally evaluate the stability of \TheName{} along two dimensions: robustness to benchmark subsampling and run-to-run variation under repeated full evaluations.
For sample-size stability, we randomly subsample tasks from each benchmark with different random seeds and compare the resulting \TheName{} leaderboard against
the full-task leaderboard.
For run-to-run stability, we repeat the full HumanEval evaluation multiple times and measure agreement across independent runs.
Together, these analyses test whether \TheName{}'s rankings are driven by a small set of tasks or by stochastic variation in LLM judging.
\begin{table}[ht]
\centering
\small
\setlength{\tabcolsep}{1pt}
\begin{tabular}{lccc}
\toprule
Benchmark & Full$(N)$ & Agreement & Top-1 recovery \\
\midrule
HumanEval & 162(80) & $0.97$ (min $0.90$) & $100\%$ \\
GSM8K & 150(80) & $0.95$ (min $0.92$) & $100\%$ \\
MMLU & 110(80) & $0.98$ (min $0.91$) & $100\%$ \\
\bottomrule
\end{tabular}
\caption{
Sample-size $N$ stability of \TheName{}.
Subsample agreement is measured by Spearman's rank correlation $\rho$ against the full-task \TheName{} leaderboard.
}
\label{tab:app-sample-stability}
\end{table}

\noindent \textbf{Sample-Size Stability.}
Table~\ref{tab:app-sample-stability} reports the agreement between subsampled and full-task \TheName{} leaderboards.
Across HumanEval, GSM8K, and MMLU, using roughly half of the benchmark tasks is sufficient to recover
a highly similar ranking, with Spearman correlation at or above $0.95$ and minimum correlation no lower than $0.90$ across 30 random seeds.
Top-1 recovery is also stable, reaching $100\%$ on HumanEval and MMLU and GSM8K.
Overall, these results suggest that \TheName{} does not require evaluating the entire benchmark to recover the main leaderboard structure.

\noindent \textbf{Run-to-Run Stability.}
We also measure run-to-run stability on HumanEval using repeated full evaluations.
Table~\ref{tab:app-rerun-stability} summarizes the production-configuration cluster, consisting of four independent full runs.
The mean Spearman correlation across runs is $0.972$, with a minimum of $0.952$, and the top-ranked model is reproduced in all pairwise run comparisons.
This indicates that under the final production configuration, \TheName{} produces
\begin{wraptable}{r}{0.45\textwidth}
\centering
\small
\setlength{\tabcolsep}{5pt}
\begin{tabular}{cccc}
\toprule
 Runs & Mean $\rho$ & Min $\rho$ & Top-1 match \\
\midrule
 4 & $0.972$ & $0.952$ & $6/6$ ($100\%$) \\
\bottomrule
\end{tabular}
\caption{
Run-to-run stability on HumanEval.
The cluster contains four independent full runs, yielding six pairwise run comparisons.
}
\label{tab:app-rerun-stability}
\end{wraptable}
stable rankings across repeated executions rather than relying on a single lucky judge trajectory.

Taken together, the subsampling and repeated-run analyses show that \TheName{} is stable at both the task-sampling level and the execution level.
In practice, evaluating roughly half of a benchmark is already sufficient to recover the main ranking structure on the larger benchmarks, while repeated full HumanEval runs produce highly consistent leaderboards.

\subsubsection{Leaderboards and Model Candidates}
\label{app:leaderboards-models}

For each benchmark, we evaluate a fixed pool of eight candidate systems selected from the saturated top region of the corresponding benchmark setting.
All protocols receive the same cached candidate outputs for a given task, so differences in the resulting leaderboard come from the evaluation protocol rather than from resampling model generations.
Table~\ref{tab:app-candidate-pools} lists the benchmark subsets, native metrics, and candidate pools used in the experiments.

The detailed per-benchmark leaderboards are shown in Table~\ref{tab:app-all-leaderboards}.
Each cell reports the model's rank and score under a protocol.
The \TheName{} column corresponds to the adaptive tournament protocol, while \textsc{FullPair} denotes exhaustive pairwise judging over all candidate pairs.
The top-1 agreement analysis directly explains how the selected candidate pools affect the most visible leaderboard decision.
Table~\ref{tab:app-top1-agreement} reports whether each protocol recovers the same top-ranked model as \textsc{FullPair}.
This top-1 criterion complements rank-correlation metrics by focusing on the single winner that would appear at the top of a benchmark leaderboard.
\TheName{} achieves $4/4$ top-1 agreement with \textsc{FullPair}, matching the exhaustive pairwise winner on HumanEval, BFCL-v2, GSM8K, and MMLU.

\begin{table}[h]
\centering
\small
\setlength{\tabcolsep}{1.5pt}
\begin{tabular}{lccccc}
\toprule
Protocol & HumanEval & BFCL & GSM8K & MMLU & Total \\
\midrule
Original  & -- & -- & -- & -- & 0/4 \\
Pointwise & \checkmark & \checkmark & -- & \checkmark & 3/4 \\
FixedRub  & \checkmark & \checkmark & -- & -- & 2/4 \\
Listwise  & \checkmark & -- & -- & \checkmark & 2/4 \\
FlatBrk   & \checkmark & \checkmark & -- & \checkmark & 3/4 \\
\rowcolor{gray!40}
\TheName{} & \checkmark & \checkmark & \checkmark & \checkmark & \textbf{4/4} \\
\bottomrule
\end{tabular}
\caption{Top-1 agreement with exhaustive pairwise judging. A checkmark means that the protocol selects the same top-ranked candidate as \textsc{FullPair} on that benchmark.}
\label{tab:app-top1-agreement}
\end{table}

\subsubsection{API Calls, Token Consumption, and Estimated Cost}
\label{app:dollar-cost}

We report (i) the theoretical per-task LLM-call count, (ii) the per-call
input / output token consumption, and (iii) the end-to-end
dollar cost. 

\noindent \textbf{Theoretical Per-Task Call Count.}
As summarized in Table~\ref{tab:app-theoretical-cost}, for $N=8$ candidates,
\textsc{FullPair} schedules $N(N{-}1)/2 = 28$ pairwise calls per task,
\textsc{FlatBrk} and \TheName{} schedule $1$ seeding $+ (N{-}1) = 8$
calls per task in the bracket, \TheName{} additionally issues
${\sim}3.9$ evolution / local-critique calls per task. Listwise needs only
one rank-$N$ call per task. 

\begin{wraptable}{r}{0.45\textwidth}
\centering
\setlength{\tabcolsep}{2pt}
\small
\begin{tabular}{lcc}
\toprule
Method & Calls / task & Asymptotic cost \\
\midrule
Original  & 0          & $O(1)$ \\
Listwise  & 1          & $O(1)$ \\
Pointwise & 8          & $O(N)$ \\
FixedRub  & $8 + 1/M$  & $O(N)$ \\
FlatBrk   & 8          & $O(N)$ \\
\rowcolor{gray!40}
\TheName{} & ${\sim}11.9$ & $O(N) +$ evolution \\
FullPair  & 28         & $O(N^2)$ \\
\bottomrule
\end{tabular}
\caption{Theoretical per-task API-call cost for $N=8$ candidates. \textsc{FixedRub}'s
rubric-generation call is amortised over $M$ tasks. \TheName{}'s extra
${\sim}3.9$ calls per task are checklist-evolution operators plus local critique in semi-final / final
matches.}
\label{tab:app-theoretical-cost}
\end{wraptable}

\noindent \textbf{Token Consumption.}
We report token consumption using the actual completions from
the \textsc{HumanEval} runs. For each protocol, we count both input tokens
(system prompts, task prompts, principles/checklists, and candidate outputs)
and output tokens, including visible responses and billed reasoning tokens.
All methods are evaluated under the same accounting rule, so the relative
comparison is consistent across protocols.

It is worth noting that although \textsc{Listwise} makes only one call per task, each call ranks all $N=8$ candidates jointly and therefore still consumes a
non-trivial number of tokens. In contrast, \textsc{FullPair} is expensive
because it exhaustively compares all candidate pairs, leading to both many
calls and high total token consumption. \TheName{} reduces this cost by
using a selective comparison structure while avoiding exhaustive all-pair
judging.


\noindent \textbf{Estimated Cost.}
We further convert token consumption into monetary cost using the same
GPT-5-tier pricing proxy for all protocols: \$2.00 per 1M input tokens and
\$17.00 per 1M output tokens. We do not apply prompt caching, so the estimates should be interpreted as conservative
end-to-end costs.

As shown in Table~\ref{tab:app-cost-estimation}, \TheName{} substantially
reduces the cost compared with exhaustive pairwise evaluation. On the full
\textsc{HumanEval} benchmark, \TheName{} costs \$58.67, compared with
\$173.79 for \textsc{FullPair}, corresponding to a \$115.1 reduction
($66.2\%$ lower cost). The same trend holds after linear extrapolation to
$1{,}000$ candidate-evaluation tasks, where \TheName{} costs only
$0.34\times$ as much as \textsc{FullPair}.


\begin{table}[t]
\centering
\small
\captionsetup[subtable]{justification=raggedright,singlelinecheck=false}

\begin{subtable}[t]{0.48\textwidth}
\vspace{0pt}
\centering
\setlength{\tabcolsep}{3pt}
\begin{tabular*}{\linewidth}{@{\extracolsep{\fill}}lrrrr@{}}
\toprule
Protocol & Calls & Input & Output & Total \\
\midrule
\textsc{Listwise}  &    162 &  0.69 & 0.32 &  1.00 \\
\textsc{Pointwise} & 1{,}296 &  4.36 & 0.49 &  4.86 \\
\textsc{FixedRub}  & 1{,}297 &  2.81 & 1.17 &  3.98 \\
\textsc{FlatBrk}   & 1{,}296 &  8.60 & 1.79 & 10.39 \\
\rowcolor{gray!25}
\TheName{}         & 1{,}926 & 10.63 & 2.20 & 12.83 \\
\textsc{FullPair}  & 4{,}536 & 30.99 & 6.58 & 37.57 \\
\bottomrule
\end{tabular*}
\caption{End-to-end token consumption on \textsc{HumanEval}. Token counts are reported in millions.}
\label{tab:app-token-consumption}
\end{subtable}
\hfill
\begin{subtable}[t]{0.48\textwidth}
\vspace{0pt}
\centering
\setlength{\tabcolsep}{3pt}
\begin{tabular*}{\linewidth}{@{\extracolsep{\fill}}lrrr@{}}
\toprule
Protocol & Cost & Cost / 1k & Rel. \\
\midrule
\textsc{Listwise}  &   6.74 &    42 & 0.04$\times$ \\
\textsc{Pointwise} &  17.10 &   106 & 0.10$\times$ \\
\textsc{FixedRub}  &  25.46 &   157 & 0.15$\times$ \\
\textsc{FlatBrk}   &  47.62 &   294 & 0.27$\times$ \\
\rowcolor{gray!25}
\TheName{}         &  58.67 &   362 & 0.34$\times$ \\
\textsc{FullPair}  & 173.79 & 1{,}073 & 1.00$\times$ \\
\bottomrule
\end{tabular*}
\caption{Estimated cost on \textsc{HumanEval}. Costs are in USD. ``Rel.'' is relative to \textsc{FullPair}.}
\label{tab:app-cost-estimation}
\end{subtable}
\caption{Efficiency comparison on \textsc{HumanEval} with $N=8$ candidate models.
\TheName{} uses only $34.2\%$ of the tokens and $33.8\%$ of the cost required by exhaustive \textsc{FullPair}.}
\label{tab:app-efficiency}
\end{table}

\section{Related Work}
Language-model evaluation has evolved from shared benchmark suites to broader and more dynamic ecosystems~\citep{wang-etal-2018-glue,wang2019superglue,srivastava2023bigbench,liang2023helm,kiela-etal-2021-dynabench}. 
However, strong models increasingly saturate native metrics, motivating benchmark-reuse methods such as metric reweighting~\citep{barbosasilva2022benchmarksaturation,akhtar2026benchmarksplateau,etzine2025revitalizing}. 
Meanwhile, LLM-as-a-judge evaluation, including MT-Bench, Chatbot Arena, G-Eval, and Prometheus, enables scalable open-ended assessment but remains vulnerable to position, self-enhancement, verbosity, and length biases~\citep{zheng2023llmjudge,chiang2024chatbotarena,liu2023geval,kim2024prometheus,wang2023llmfair,dubois2024lengthcontrolledalpacaeval}. 
Pairwise ranking can expose subtle differences in saturated regimes, yet exhaustive comparison is quadratic; prior work shows that sparse or actively selected comparisons can recover rankings more efficiently~\citep{jamieson2011activeranking,wauthier2013efficientranking}. 
\TheName{} therefore reuses fixed benchmark tasks and candidate outputs, combining seeded near-linear pairwise tournaments with adaptive checklist refinement; unlike Meta-Rewarding, our meta-judge refines evaluation criteria rather than training reward models or directly selecting winners~\citep{wu2025metarewarding}.

\section{Conclusion}
We present \TheName{}, a seeded-elimination evaluation protocol for recovering ranking signal from saturated benchmarks without constructing new tasks or changing the candidate outputs. 
By combining coarse seeding, pairwise principle-level judging, and adaptive checklist refinement by an LLM-as-a-Meta-Judge, \TheName{} increases evaluation resolution where strong candidates are difficult to separate while preserving stable aggregation through fixed task-level principles. 
Across various task settings, \TheName{} closely matches exhaustive pairwise judging and substantially reduces judge-call cost relative to all-pairs comparison. 
These results suggest that benchmark saturation is not only a property of the task set, but also of the evaluation protocol: existing saturated benchmarks can still provide useful ranking signal when evaluated with sufficiently adaptive and structured judging.

\section{Limitations}
Although \TheName{} reduces the cost of exhaustive pairwise judging, it remains dependent on the reliability of the underlying LLM judge, and structured voting cannot fully eliminate judge bias, prompt sensitivity, or occasional reasoning errors.
The seeded single-elimination design also trades completeness for efficiency: because not all candidate pairs are compared, early bracket mistakes may propagate into the final ranking.
In addition, adaptive checklist evolution adds sequential dependencies across rounds, making \TheName{} slower than one-shot listwise or pointwise baselines even though it is cheaper than full pairwise comparison.
While these limitations remain, we mitigate them where possible through fixed task-level principles, structured pairwise voting, mechanical aggregation, seeded brackets, and comparison against an exhaustive pairwise reference.

\bibliographystyle{plainnat}
\bibliography{main}

\begin{thebibliography}{21}
\providecommand{\natexlab}[1]{#1}
\providecommand{\url}[1]{\texttt{#1}}
\expandafter\ifx\csname urlstyle\endcsname\relax
  \providecommand{\doi}[1]{doi: #1}\else
  \providecommand{\doi}{doi: \begingroup \urlstyle{rm}\Url}\fi

\bibitem[Akhtar et~al.(2026)Akhtar, Reuel, Soni, Ahuja, Ammanamanchi, Rawal, Zouhar, Yadav, Whitehouse, Ki, Mickel, Choshen, {\v{S}}uppa, Batzner, Chim, Sania, Long, Rahmani, Knight, Nan, Raj, Fan, Singh, Sahoo, Habba, Gohar, Pawar, Scholz, Subramonian, Ni, Kochenderfer, Koyejo, Sachan, Biderman, Talat, Ghosh, and Solaiman]{akhtar2026benchmarksplateau}
Mubashara Akhtar, Anka Reuel, Prajna Soni, Sanchit Ahuja, Pawan~Sasanka Ammanamanchi, Ruchit Rawal, Vil{\'e}m Zouhar, Srishti Yadav, Chenxi Whitehouse, Dayeon Ki, Jennifer Mickel, Leshem Choshen, Marek {\v{S}}uppa, Jan Batzner, Jenny Chim, Jeba Sania, Yanan Long, Hossein~A. Rahmani, Christina Knight, Yiyang Nan, Jyoutir Raj, Yu~Fan, Shubham Singh, Subramanyam Sahoo, Eliya Habba, Usman Gohar, Siddhesh Pawar, Robert Scholz, Arjun Subramonian, Jingwei Ni, Mykel Kochenderfer, Sanmi Koyejo, Mrinmaya Sachan, Stella Biderman, Zeerak Talat, Avijit Ghosh, and Irene Solaiman.
\newblock When {AI} benchmarks plateau: A systematic study of benchmark saturation.
\newblock \emph{Computing Research Repository}, arXiv:2602.16763, 2026.
\newblock URL \url{https://arxiv.org/abs/2602.16763v1}.
\newblock Version 1.

\bibitem[Barbosa-Silva et~al.(2022)Barbosa-Silva, Ott, Blagec, Brauner, and Samwald]{barbosasilva2022benchmarksaturation}
Adriano Barbosa-Silva, Simon Ott, Kathrin Blagec, Jan~M. Brauner, and Matthias Samwald.
\newblock Mapping global dynamics of benchmark creation and saturation in artificial intelligence.
\newblock \emph{Computing Research Repository}, arXiv:2203.04592, 2022.
\newblock URL \url{https://dblp.org/rec/journals/corr/abs-2203-04592}.

\bibitem[Chen et~al.(2021)Chen, Tworek, Jun, Yuan, Ponde~de Oliveira~Pinto, Kaplan, Edwards, Burda, Joseph, Brockman, Ray, Puri, Krueger, Petrov, Khlaaf, Sastry, Mishkin, Chan, Gray, Ryder, Pavlov, Power, Kaiser, Bavarian, Winter, Tillet, Such, Cummings, Plappert, Chantzis, Barnes, Herbert-Voss, Guss, Nichol, Paino, Tezak, Tang, Babuschkin, Balaji, Jain, Saunders, Hesse, Carr, Leike, Achiam, Misra, Morikawa, Radford, Knight, Brundage, Murati, Mayer, Welinder, McGrew, Amodei, McCandlish, Sutskever, and Zaremba]{chen2021evaluatingcode}
Mark Chen, Jerry Tworek, Heewoo Jun, Qiming Yuan, Henrique Ponde~de Oliveira~Pinto, Jared Kaplan, Harri Edwards, Yuri Burda, Nicholas Joseph, Greg Brockman, Alex Ray, Raul Puri, Gretchen Krueger, Michael Petrov, Heidy Khlaaf, Girish Sastry, Pamela Mishkin, Brooke Chan, Scott Gray, Nick Ryder, Mikhail Pavlov, Alethea Power, Lukasz Kaiser, Mohammad Bavarian, Clemens Winter, Philippe Tillet, Felipe~Petroski Such, Dave Cummings, Matthias Plappert, Fotios Chantzis, Elizabeth Barnes, Ariel Herbert-Voss, William~Hebgen Guss, Alex Nichol, Alex Paino, Nikolas Tezak, Jie Tang, Igor Babuschkin, Suchir Balaji, Shantanu Jain, William Saunders, Christopher Hesse, Andrew~N. Carr, Jan Leike, Josh Achiam, Vedant Misra, Evan Morikawa, Alec Radford, Matthew Knight, Miles Brundage, Mira Murati, Katie Mayer, Peter Welinder, Bob McGrew, Dario Amodei, Sam McCandlish, Ilya Sutskever, and Wojciech Zaremba.
\newblock Evaluating large language models trained on code.
\newblock \emph{Computing Research Repository}, arXiv:2107.03374, 2021.
\newblock URL \url{https://arxiv.org/abs/2107.03374}.

\bibitem[Chiang et~al.(2024)Chiang, Zheng, Sheng, Angelopoulos, Li, Li, Zhang, Zhu, Jordan, Gonzalez, and Stoica]{chiang2024chatbotarena}
Wei-Lin Chiang, Lianmin Zheng, Ying Sheng, Anastasios~Nikolas Angelopoulos, Tianle Li, Dacheng Li, Hao Zhang, Banghua Zhu, Michael~I. Jordan, Joseph~E. Gonzalez, and Ion Stoica.
\newblock Chatbot arena: An open platform for evaluating {LLM}s by human preference.
\newblock In \emph{International Conference on Machine Learning}, 2024.
\newblock URL \url{https://openreview.net/forum?id=3MW8GKNyzI}.

\bibitem[Cobbe et~al.(2021)Cobbe, Kosaraju, Bavarian, Chen, Jun, Kaiser, Plappert, Tworek, Hilton, Nakano, Hesse, and Schulman]{cobbe2021trainingverifiers}
Karl Cobbe, Vineet Kosaraju, Mohammad Bavarian, Mark Chen, Heewoo Jun, Lukasz Kaiser, Matthias Plappert, Jerry Tworek, Jacob Hilton, Reiichiro Nakano, Christopher Hesse, and John Schulman.
\newblock Training verifiers to solve math word problems.
\newblock \emph{Computing Research Repository}, arXiv:2110.14168, 2021.
\newblock URL \url{https://arxiv.org/abs/2110.14168}.

\bibitem[Dubois et~al.(2024)Dubois, Galambosi, Liang, and Hashimoto]{dubois2024lengthcontrolledalpacaeval}
Yann Dubois, Balazs Galambosi, Percy Liang, and Tatsunori~B. Hashimoto.
\newblock Length-controlled {AlpacaEval}: A simple way to debias automatic evaluators.
\newblock \emph{Computing Research Repository}, arXiv:2404.04475, 2024.
\newblock URL \url{https://arxiv.org/abs/2404.04475}.

\bibitem[Etzine et~al.(2025)Etzine, Mazzawi, and Wolf]{etzine2025revitalizing}
Ben Etzine, Hanna Mazzawi, and Lior Wolf.
\newblock Revitalizing saturated benchmarks: A weighted metric approach for enhanced model differentiation.
\newblock \emph{Computing Research Repository}, arXiv:2503.05551, 2025.
\newblock URL \url{https://dblp.org/rec/journals/corr/abs-2503-05551}.

\bibitem[Hendrycks et~al.(2021)Hendrycks, Burns, Basart, Zou, Mazeika, Song, and Steinhardt]{hendrycks2021mmlu}
Dan Hendrycks, Collin Burns, Steven Basart, Andy Zou, Mantas Mazeika, Dawn Song, and Jacob Steinhardt.
\newblock Measuring massive multitask language understanding.
\newblock In \emph{International Conference on Learning Representations}, 2021.
\newblock URL \url{https://openreview.net/forum?id=d7KBjmI3GmQ}.

\bibitem[Jamieson and Nowak(2011)]{jamieson2011activeranking}
Kevin~G. Jamieson and Robert~D. Nowak.
\newblock Active ranking using pairwise comparisons.
\newblock In \emph{Advances in Neural Information Processing Systems}, volume~24, 2011.
\newblock URL \url{https://proceedings.neurips.cc/paper/2011/hash/6c14da109e294d1e8155be8aa4b1ce8e-Abstract.html}.

\bibitem[Kiela et~al.(2021)Kiela, Bartolo, Nie, Kaushik, Geiger, Wu, Vidgen, Prasad, Singh, Ringshia, Ma, Thrush, Riedel, Waseem, Stenetorp, Jia, Bansal, Potts, and Williams]{kiela-etal-2021-dynabench}
Douwe Kiela, Max Bartolo, Yixin Nie, Divyansh Kaushik, Atticus Geiger, Zhengxuan Wu, Bertie Vidgen, Grusha Prasad, Amanpreet Singh, Pratik Ringshia, Zhiyi Ma, Tristan Thrush, Sebastian Riedel, Zeerak Waseem, Pontus Stenetorp, Robin Jia, Mohit Bansal, Christopher Potts, and Adina Williams.
\newblock {Dynabench}: Rethinking benchmarking in {NLP}.
\newblock In \emph{Proceedings of the 2021 Conference of the North American Chapter of the Association for Computational Linguistics: Human Language Technologies}, pages 4110--4124, Online, 2021. Association for Computational Linguistics.
\newblock URL \url{https://aclanthology.org/2021.naacl-main.324}.

\bibitem[Kim et~al.(2024)Kim, Suk, Longpre, Yoon, Kim, Lee, Yun, Shin, Kim, Thorne, and Seo]{kim2024prometheus}
Seungone Kim, Jamin Suk, Shayne Longpre, Bill Yoon, Jamin~Shin Kim, Jiyoung Lee, Sangdoo Yun, Seongil Shin, Sungdong Kim, James Thorne, and Minjoon Seo.
\newblock Prometheus: Inducing fine-grained evaluation capability in language models.
\newblock In \emph{International Conference on Learning Representations}, 2024.
\newblock URL \url{https://openreview.net/forum?id=8euJaTveKw}.

\bibitem[Liang et~al.(2023)Liang, Bommasani, Lee, Tsipras, Soylu, Yasunaga, Zhang, Narayanan, Wu, Kumar, Newman, Yuan, Yan, Zhang, Cosgrove, Manning, R{\'e}, Acosta-Navas, Hudson, Zelikman, Durmus, Ladhak, Rong, Ren, Yao, Wang, Santhanam, Orr, Zheng, Yuksekgonul, Suzgun, Kim, Guha, Chatterji, Khattab, Henderson, Huang, Chi, Xie, Santurkar, Ganguli, Hashimoto, Icard, Zhang, Chaudhary, Wang, Li, Mai, Zhang, and Koreeda]{liang2023helm}
Percy Liang, Rishi Bommasani, Tony Lee, Dimitris Tsipras, Dilara Soylu, Michihiro Yasunaga, Yian Zhang, Deepak Narayanan, Yuhuai Wu, Ananya Kumar, Benjamin Newman, Binhang Yuan, Bobby Yan, Ce~Zhang, Christopher Cosgrove, Christopher~D. Manning, Christopher R{\'e}, Diana Acosta-Navas, Drew~A. Hudson, Eric Zelikman, Esin Durmus, Faisal Ladhak, Frieda Rong, Hongyu Ren, Huaxiu Yao, Jue Wang, Keshav Santhanam, Laurel Orr, Lucia Zheng, Mert Yuksekgonul, Mirac Suzgun, Nathan Kim, Neel Guha, Niladri Chatterji, Omar Khattab, Peter Henderson, Qian Huang, Ryan Chi, Sang~Michael Xie, Shibani Santurkar, Surya Ganguli, Tatsunori Hashimoto, Thomas Icard, Tianyi Zhang, Vishrav Chaudhary, William Wang, Xuechen Li, Yifan Mai, Yuhui Zhang, and Yuta Koreeda.
\newblock Holistic evaluation of language models.
\newblock \emph{Transactions on Machine Learning Research}, 2023.
\newblock URL \url{https://openreview.net/forum?id=iO4LZibEqW}.

\bibitem[Liu et~al.(2023)Liu, Iter, Xu, Wang, Xu, and Zhu]{liu2023geval}
Yang Liu, Dan Iter, Yichong Xu, Shuohang Wang, Ruochen Xu, and Chenguang Zhu.
\newblock {G-Eval}: {NLG} evaluation using {GPT}-4 with better human alignment.
\newblock In \emph{Proceedings of the 2023 Conference on Empirical Methods in Natural Language Processing}, pages 2511--2522. Association for Computational Linguistics, 2023.
\newblock \doi{10.18653/v1/2023.emnlp-main.153}.
\newblock URL \url{https://aclanthology.org/2023.emnlp-main.153}.

\bibitem[Patil et~al.(2025)Patil, Mao, Yan, Jamba, Shankar, Gonzalez, and Stoica]{patil2025bfcl}
Shishir~G. Patil, Tianjun Mao, Xupeng Yan, Siva~E. Jamba, S.~Shankar, Joseph~E. Gonzalez, and Ion Stoica.
\newblock The berkeley function calling leaderboard.
\newblock In \emph{International Conference on Machine Learning}, 2025.
\newblock URL \url{https://proceedings.mlr.press/v267/patil25a.html}.

\bibitem[Srivastava et~al.(2023)Srivastava, Rastogi, Rao, Shoeb, Abid, Fisch, Brown, Santoro, Gupta, Garriga-Alonso, Kluska, Lewkowycz, Agarwal, Power, and Others]{srivastava2023bigbench}
Aarohi Srivastava, Abhinav Rastogi, Abhishek Rao, Abu Awal~Md Shoeb, Abubakar Abid, Adam Fisch, Adam~R. Brown, Adam Santoro, Aditya Gupta, Adri{\`a} Garriga-Alonso, Agnieszka Kluska, Aitor Lewkowycz, Akshat Agarwal, Alethea Power, and Others.
\newblock Beyond the imitation game: Quantifying and extrapolating the capabilities of language models.
\newblock \emph{Transactions on Machine Learning Research}, 2023.
\newblock URL \url{https://dblp.org/rec/journals/tmlr/SrivastavaRRSAF23}.

\bibitem[Wang et~al.(2018)Wang, Singh, Michael, Hill, Levy, and Bowman]{wang-etal-2018-glue}
Alex Wang, Amanpreet Singh, Julian Michael, Felix Hill, Omer Levy, and Samuel~R. Bowman.
\newblock {GLUE}: A multi-task benchmark and analysis platform for natural language understanding.
\newblock In \emph{Proceedings of the 2018 {EMNLP} Workshop {B}lackbox{NLP}: Analyzing and Interpreting Neural Networks for {NLP}}, pages 353--355, Brussels, Belgium, 2018. Association for Computational Linguistics.
\newblock URL \url{https://aclanthology.org/W18-5446}.

\bibitem[Wang et~al.(2019)Wang, Pruksachatkun, Nangia, Singh, Michael, Hill, Levy, and Bowman]{wang2019superglue}
Alex Wang, Yada Pruksachatkun, Nikita Nangia, Amanpreet Singh, Julian Michael, Felix Hill, Omer Levy, and Samuel~R. Bowman.
\newblock {SuperGLUE}: A stickier benchmark for general-purpose language understanding systems.
\newblock In \emph{Advances in Neural Information Processing Systems}, volume~32, 2019.
\newblock URL \url{https://papers.nips.cc/paper/8589-superglue-a-stickier-benchmark-for-general-purpose-language-understanding-systems}.

\bibitem[Wang et~al.(2023)Wang, Li, Chen, Cai, Zhu, Lin, Cao, Liu, Liu, and Sui]{wang2023llmfair}
Peiyi Wang, Lei Li, Liang Chen, Zefan Cai, Dawei Zhu, Binghuai Lin, Yunbo Cao, Lingpeng Liu, Tianyu Liu, and Zhifang Sui.
\newblock Large language models are not fair evaluators.
\newblock \emph{Computing Research Repository}, arXiv:2305.17926, 2023.
\newblock URL \url{https://arxiv.org/abs/2305.17926}.

\bibitem[Wauthier et~al.(2013)Wauthier, Jojic, and Jordan]{wauthier2013efficientranking}
Fabian~L. Wauthier, Nebojsa Jojic, and Michael~I. Jordan.
\newblock Efficient ranking from pairwise comparisons.
\newblock In \emph{International Conference on Machine Learning}, pages 109--117, 2013.
\newblock URL \url{https://proceedings.mlr.press/v28/wauthier13.html}.

\bibitem[Wu et~al.(2025)Wu, Yuan, Golovneva, Xu, Tian, Jiao, Weston, and Sukhbaatar]{wu2025metarewarding}
Tianhao Wu, Weizhe Yuan, Olga Golovneva, Jing Xu, Yuandong Tian, Jian Jiao, Jason Weston, and Sainbayar Sukhbaatar.
\newblock Meta-rewarding language models: Self-improving alignment with {LLM}-as-a-meta-judge.
\newblock In \emph{Proceedings of the 2025 Conference on Empirical Methods in Natural Language Processing}, pages 11537--11554. Association for Computational Linguistics, 2025.
\newblock URL \url{https://aclanthology.org/2025.emnlp-main.704/}.

\bibitem[Zheng et~al.(2023)Zheng, Chiang, Sheng, Zhuang, Wu, Zhuang, Lin, Li, Li, Xing, Zhang, Gonzalez, and Stoica]{zheng2023llmjudge}
Lianmin Zheng, Wei-Lin Chiang, Ying Sheng, Siyuan Zhuang, Zhanghao Wu, Yonghao Zhuang, Zi~Lin, Zhuohan Li, Dacheng Li, Eric~P. Xing, Hao Zhang, Joseph~E. Gonzalez, and Ion Stoica.
\newblock Judging {LLM}-as-a-judge with {MT}-bench and chatbot arena.
\newblock In \emph{Advances in Neural Information Processing Systems}, volume~36, 2023.
\newblock URL \url{https://proceedings.neurips.cc/paper_files/paper/2023/hash/91f18a1287b398d378ef22505bf41832-Abstract-Datasets_and_Benchmarks.html}.

\end{thebibliography}

\clearpage
\appendix


\section{Detailed Experimental Settings}
\label{app:prompts}

This section provides implementation details for the evaluation protocol.
We first describe the predefined task-level principles and seed checklists used to initialize all runs.
We then summarize the judge and evolution model choices and provide the main prompt templates used for seeding, pairwise judging, and checklist evolution.

\subsection{Predefined Principles and Seed Checklists}
\label{app:rubric}

\TheName{} starts from a fixed two-layer rubric for each task type.
The outer layer consists of task-level principles, which define broad quality dimensions such as correctness, robustness, efficiency, factuality, or schema compliance.
Each principle has a positive weight and the weights within a task type sum to $1.0$.
The inner layer is the seed checklist $\Gamma_0$, which contains one checklist item per principle.
Each seed item operationalizes its matched principle as a concrete judging probe used in the initial pairwise comparisons.
All protocols use the same predefined principles; \textsc{FlatBrk} and \textsc{FullPair} also keep the same seed checklist fixed throughout evaluation, while \TheName{} starts from this checklist and then adaptively adds finer-grained checklist items in later rounds.

\begin{table}[ht]
\centering
\small
\setlength{\tabcolsep}{2pt}
\renewcommand{\arraystretch}{1.1}
\begin{tabular}{lccc}
\toprule
Benchmark & Task type & \# Principles & \# Seed items \\
\midrule
HumanEval & \texttt{code\_generation} & 6 & 6 \\
GSM8K & \texttt{math\_reasoning} & 4 & 4 \\
MMLU & \texttt{general\_qa} & 5 & 5 \\
BFCL-v2 & \texttt{tool\_calling} & 5 & 5 \\
\bottomrule
\end{tabular}
\caption{Predefined principle and seed-checklist sizes. Each task type has one seed checklist item per principle.}
\label{tab:app-rubric-overview}
\end{table}

\paragraph{Code Generation.}
For HumanEval, the predefined principles cover functional correctness, interface and integration, edge-case robustness, algorithmic efficiency, code quality and maintainability, and safety or side effects.
The corresponding seed checklist asks whether the implementation handles untested semantic details, preserves the required callable interface, covers valid edge cases, uses appropriate asymptotic complexity, remains readable and idiomatic, and avoids unintended mutation, I/O, global state, or unsafe operations.

\paragraph{Mathematical Reasoning.}
For GSM8K, the principles are correctness, reasoning soundness, completeness, and clarity.
The seed checklist tests whether the final numeric answer is correct in the requested form, whether every algebraic or arithmetic step is locally justified, whether all required sub-questions or cases are addressed, and whether the final answer is unambiguous and easy to identify.

\paragraph{General Question Answering.}
For MMLU, the principles are goal achievement, factual accuracy, helpfulness, safety, and style.
The seed checklist asks whether the response directly answers the specific question, whether factual claims are accurate and uncertainty is calibrated, whether the answer is actionable for the user’s situation, whether it avoids harmful or deceptive content, and whether its length and structure fit the question.

\paragraph{Tool Calling.}
For BFCL-v2, the principles are function selection, argument correctness, schema compliance, irrelevance handling, and response form.
The seed checklist evaluates whether the model calls exactly the right function or functions, provides accurate and well-typed arguments, follows the declared schema, correctly declines when no available function applies, and emits either a clean tool call or a concise refusal/clarification in the expected form.


\subsection{Judge/Evolving Model Choices}
For \TheName{}, both the judge model and the evolution model are \texttt{gpt-5.5}; the same judge model is also used for all LLM-based baselines.

\subsection{Seeding Prompt}
The seeding prompt is used once per task before the tournament begins.
Its system prompt defines the tiering objective and instructs the judge to group semantically similar candidates together, while its user prompt supplies the task, evaluation principles, and candidate outputs.

\begin{tcolorbox}[
  title={Seeding Prompt: System Message},
  breakable,
  colback=gray!3,
  colframe=gray!45,
  fonttitle=\bfseries,
  left=1mm,
  right=1mm,
  top=1mm,
  bottom=1mm
]
You are an expert evaluator. Given N candidate responses to the same task,
roughly rank them into k tiers, where Tier 1 is best and Tier k is worst.

Judge only substantive task-relevant quality differences. Do not tier by
cosmetic formatting, response style, or superficial wording. Responses that are
semantically equivalent should be placed in the same tier.

Use the provided evaluation principles as guidance. You do not need to produce
an exact ranking inside each tier; focus on which candidates are clearly
stronger or weaker.

Return only valid JSON:

\{
  "tiers": \{
    "1": ["model\_A", "model\_B"],
    "2": ["model\_C"]
  \},
  "reasoning": "brief explanation"
\}

Every model\_id provided must appear in exactly one tier.
\end{tcolorbox}

The corresponding user message instantiates the generic tiering instruction with a concrete benchmark task, the fixed task-level principles, and the candidate responses to be seeded.

\begin{tcolorbox}[
  title={Seeding Prompt: User Message},
  breakable,
  colback=gray!3,
  colframe=gray!45,
  fonttitle=\bfseries,
  left=1mm,
  right=1mm,
  top=1mm,
  bottom=1mm
]
Task:

\$\{task\_prompt\}

Evaluation Principles:

\$\{principles\_block\}

Candidate responses:

\$\{solutions\_block\}

Assign each candidate to one of \$\{k\} tiers.
\end{tcolorbox}

\subsection{Pairwise Judge Prompt}
The pairwise judge prompt is the core comparison prompt used by \TheName{}, \textsc{FlatBrk}, and \textsc{FullPair}.
The system prompt fixes the voting protocol: the judge must vote left, right, or tie for each principle and checklist item, and the final verdict is derived mechanically from confidence-weighted principle votes.

\begin{tcolorbox}[
  title={Pairwise Judge Prompt: System Message},
  breakable,
  colback=gray!3,
  colframe=gray!45,
  fonttitle=\bfseries,
  left=1mm,
  right=1mm,
  top=1mm,
  bottom=1mm
]
You are an expert evaluator performing a pairwise comparison of two candidate
responses to the same task. Your job is to decide, for each Principle and each
Checklist item, whether the left response is better, the right response is
better, or they are tied.

Judge only against the requirements stated in the task. Do not invent extra
requirements. Do not reward or penalize cosmetic differences.

For each Principle, output:
- vote: "left", "right", or "tie"
- confidence: a number from 0.0 to 1.0
- reasoning: one sentence citing the observable behavior that decided the vote

Use "tie" when there is no concrete, non-cosmetic difference. When a real
difference exists, vote for the better side and cite the specific behavior.

The overall verdict must be mechanical. Compute the confidence-weighted vote
sum over Principles:
- left vote = -1
- right vote = +1
- tie = 0

The verdict is:
- "left" if the weighted sum is negative
- "right" if the weighted sum is positive
- "tie" if the weighted sum is effectively zero

Do not use an unconstrained holistic impression.

For each Checklist item, apply the same pairwise voting rule:
vote "left", "right", or "tie", with confidence.

Return only valid JSON:

\{
  "verdict": "left" | "right" | "tie",
  "principle\_scores": [
    \{
      "principle\_id": "P1",
      "vote": "left" | "right" | "tie",
      "confidence": <0-1>,
      "reasoning": "<one sentence citing the deciding behavior>"
    \}
  ],
  "checklist\_scores": [
    \{
      "item\_id": "<id>",
      "vote": "left" | "right" | "tie",
      "confidence": <0-1>
    \}
  ]
\}

You must output one vote for every Principle and every Checklist item.
\end{tcolorbox}

The user message supplies the concrete task, the fixed principles, the current checklist, and the two candidate responses being compared.
\TheName{} differs from \textsc{FlatBrk} and \textsc{FullPair} only in how matches are scheduled and how the checklist is updated; the pairwise judge format remains shared.

\begin{tcolorbox}[
  title={Pairwise Judge Prompt: User Message},
  breakable,
  colback=gray!3,
  colframe=gray!45,
  fonttitle=\bfseries,
  left=1mm,
  right=1mm,
  top=1mm,
  bottom=1mm
]
Task:

\$\{task\_prompt\}

Evaluation Principles:

\$\{principles\_block\}

Checklist:

\$\{checklist\_block\}

=== Response A (\$\{left\_model\}) ===

\$\{left\_output\}

=== Response B (\$\{right\_model\}) ===

\$\{right\_output\}

Decide, for each Principle and Checklist item, whether Response A or Response B
is better, or whether they are equivalent. Output only the JSON object.
\end{tcolorbox}

\subsection{Checklist-Evolution Prompt}
The checklist-evolution prompt is used only by \TheName{}.
It introduces one new adversarial checklist item after a close comparison, where a strong response loses to a slightly better response.
The system prompt constrains the new item to be non-redundant, principle-grounded, anchor-calibrated, and supported by a demonstration pair.

\begin{tcolorbox}[
  title={Checklist-Evolution System Prompt},
  breakable,
  colback=gray!3,
  colframe=gray!45,
  fonttitle=\bfseries,
  left=1mm,
  right=1mm,
  top=1mm,
  bottom=1mm
]
You are an expert rubric designer. Given a comparison where a strong response
lost to a slightly better response, generate exactly one new checklist criterion
that captures the subtle, observable quality difference that decided the match.

Rules:

1. Anti-redundancy:
   Do not generate a criterion that duplicates or paraphrases an existing
   checklist item.

2. Coverage:
   Prefer principles that currently have fewer checklist items, unless the
   distinguishing behavior clearly belongs to a more covered principle.

3. Anchor calibration:
   The new criterion must map to one fixed evaluation principle and distinguish
   between two adjacent anchor levels, such as 5 vs. 4 or 4 vs. 3.

4. Demonstration pair:
   Provide two short concrete snippets or examples:
   - higher\_snippet: illustrates the better behavior
   - lower\_snippet: illustrates the weaker behavior

5. Positive and concrete framing:
   The criterion should describe what a good response does. Avoid vague
   criteria such as "be efficient" or "follow best practice."

6. Task scope:
   Do not invent requirements outside the task description.

Return only valid JSON:

\{
  "id": "\$\{new\_id\}",
  "description": "<one concrete, anchor-grounded checklist criterion>",
  "principle\_id": "<one principle id>",
  "anchor\_split": \{
    "higher": <int 1-5>,
    "lower": <int 1-5>
  \},
  "differentiator": "<one sentence explaining the observable quality gap>",
  "demonstration\_pair": \{
    "higher\_snippet": "<short example showing the stronger behavior>",
    "lower\_snippet": "<short example showing the weaker behavior>"
  \},
  "scoring": "five\_level",
  "source": "adversarial"
\}
\end{tcolorbox}

The user message gives the evolver the task context, the losing and winning responses, the loser-side principle evidence, the current principle set, and the existing checklist.
This lets the evolver add a criterion that targets an observed unresolved gap rather than generating a generic rubric item.

\begin{tcolorbox}[
  title={Checklist-Evolution Prompt: User Message},
  breakable,
  colback=gray!3,
  colframe=gray!45,
  fonttitle=\bfseries,
  left=1mm,
  right=1mm,
  top=1mm,
  bottom=1mm
]
Task description:

\$\{task\_description\}

Loser response:

\$\{loser\_output\}

Winner response:

\$\{winner\_output\}

Per-principle evidence for the loser:

\$\{loser\_scores\}

Available principles:

\$\{principles\_block\}

Existing checklist:

\$\{existing\_items\_block\}

Items-per-principle counts:

\$\{coverage\_counts\}

Generate exactly one new checklist criterion with ID="\$\{new\_id\}" that captures
the subtle quality gap between the loser and winner without duplicating an
existing checklist item.
\end{tcolorbox}

\begin{table*}[t]
\centering
\small
\setlength{\tabcolsep}{4pt}
\begin{tabular}{p{0.12\linewidth}p{0.08\linewidth}p{0.16\linewidth}p{0.5\linewidth}}
\toprule
Benchmark & Tasks & Native metric & Candidate models \\
\midrule
HumanEval & 162 & pass@1 & \texttt{openai/gpt-5}\newline \texttt{openai/o4-mini-high}\newline \texttt{google/gemini-3.1-pro}\newline \texttt{x-ai/grok-4.20}\newline \texttt{anthropic/claude-sonnet-4.6}\newline \texttt{deepseek/deepseek-v3.2}\newline \texttt{anthropic/claude-opus-4.7}\newline \texttt{qwen/qwen3-coder-plus} \\
BFCL-v2 & 100 & success rate & \texttt{qwen/qwen3-coder}\newline \texttt{google/gemini-2.5-flash}\newline \texttt{z-ai/glm-5-turbo}\newline \texttt{google/gemini-3.1-pro}\newline \texttt{moonshotai/kimi-k2.6}\newline \texttt{openai/o4-mini-high}\newline \texttt{deepseek/deepseek-v3.2}\newline \texttt{openai/gpt-5} \\
GSM8K & 150 & numeric exact match & \texttt{anthropic/claude-sonnet-4}\newline \texttt{google/gemini-2.5-pro}\newline \texttt{openai/o4-mini-high}\newline \texttt{deepseek/deepseek-r1-0528}\newline \texttt{google/gemini-3-flash-preview}\newline \texttt{qwen/qwen3-30b-a3b-thinking-2507}\newline \texttt{openai/gpt-5}\newline \texttt{qwen/qwen3-235b-a22b} \\
MMLU & 110 & multiple-choice accuracy & \texttt{openai/gpt-5-mini}\newline \texttt{google/gemini-3.1-pro}\newline \texttt{qwen/qwen3.6-plus}\newline \texttt{openai/gpt-5.4-nano}\newline \texttt{x-ai/grok-4.3}\newline \texttt{deepseek/deepseek-chat}\newline \texttt{anthropic/claude-3.5-haiku}\newline \texttt{minimax/minimax-m2.7} \\
\bottomrule
\end{tabular}
\caption{Benchmark subsets and selected candidate models. The candidate pools are fixed across all evaluation protocols.}
\label{tab:app-candidate-pools}
\end{table*}

\begin{table*}[t]
\centering
\scriptsize
\setlength{\tabcolsep}{2.2pt}
\renewcommand{\arraystretch}{0.95}
\resizebox{\textwidth}{!}{
\begin{tabular}{llccccccc}
\toprule
Benchmark & Model & Original & Pointwise & FixedRub & Listwise & FlatBrk & FullPair & \TheName{} \\
\midrule
\multirow{8}{*}{HumanEval}
& \texttt{openai/gpt-5} & \#5 .988 & \#1 .980 & \#1 .310 & \#1 .748 & \#1 .511 & \#1 .541 & \textbf{\#1 .509} \\
& \texttt{openai/o4-mini-high} & \#3 .994 & \#3 .968 & \#7 .306 & \#2 .604 & \#2 .492 & \#2 .523 & \textbf{\#2 .488} \\
& \texttt{google/gemini-3.1-pro} & \#2 1.00 & \#4 .968 & \#8 .305 & \#4 .526 & \#3 .483 & \#4 .511 & \textbf{\#3 .486} \\
& \texttt{x-ai/grok-4.20} & \#4 .994 & \#2 .970 & \#2 .310 & \#6 .437 & \#4 .482 & \#3 .512 & \textbf{\#4 .483} \\
& \texttt{anthropic/claude-sonnet-4.6} & \#1 1.00 & \#6 .956 & \#6 .308 & \#3 .550 & \#5 .456 & \#6 .488 & \textbf{\#5 .463} \\
& \texttt{deepseek/deepseek-v3.2} & \#6 .981 & \#5 .959 & \#5 .309 & \#7 .342 & \#6 .451 & \#5 .493 & \textbf{\#6 .457} \\
& \texttt{anthropic/claude-opus-4.7} & \#7 .963 & \#8 .931 & \#4 .309 & \#5 .492 & \#8 .432 & \#7 .469 & \textbf{\#7 .440} \\
& \texttt{qwen/qwen3-coder-plus} & \#8 .963 & \#7 .932 & \#3 .309 & \#8 .301 & \#7 .437 & \#8 .461 & \textbf{\#8 .435} \\
\midrule
\multirow{8}{*}{BFCL-v2}
& \texttt{qwen/qwen3-coder} & \#3 .857 & \#1 .835 & \#1 .892 & \#2 .664 & \#1 .496 & \#1 .558 & \textbf{\#1 .532} \\
& \texttt{google/gemini-2.5-flash} & \#4 .823 & \#2 .810 & \#2 .882 & \#1 .772 & \#4 .463 & \#2 .549 & \textbf{\#2 .524} \\
& \texttt{z-ai/glm-5-turbo} & \#1 .867 & \#3 .783 & \#4 .838 & \#3 .539 & \#2 .490 & \#3 .540 & \textbf{\#3 .464} \\
& \texttt{google/gemini-3.1-pro} & \#2 .860 & \#4 .782 & \#3 .863 & \#4 .485 & \#3 .472 & \#4 .530 & \textbf{\#4 .458} \\
& \texttt{moonshotai/kimi-k2.6} & \#5 .817 & \#5 .771 & \#5 .825 & \#5 .469 & \#5 .460 & \#5 .514 & \textbf{\#5 .421} \\
& \texttt{openai/o4-mini-high} & \#7 .668 & \#6 .690 & \#7 .731 & \#6 .418 & \#7 .389 & \#7 .433 & \textbf{\#6 .368} \\
& \texttt{deepseek/deepseek-v3.2} & \#8 .653 & \#8 .676 & \#8 .729 & \#7 .365 & \#8 .383 & \#6 .438 & \textbf{\#7 .364} \\
& \texttt{openai/gpt-5} & \#6 .716 & \#7 .685 & \#6 .740 & \#8 .287 & \#6 .389 & \#8 .424 & \textbf{\#8 .354} \\
\midrule
\multirow{8}{*}{GSM8K}
& \texttt{anthropic/claude-sonnet-4} & \#6 .965 & \#4 .928 & \#4 .992 & \#2 .668 & \#4 .499 & \#1 .549 & \textbf{\#1 .503} \\
& \texttt{google/gemini-2.5-pro} & \#5 .967 & \#3 .933 & \#3 .992 & \#4 .489 & \#3 .499 & \#3 .542 & \textbf{\#2 .495} \\
& \texttt{openai/o4-mini-high} & \#4 .970 & \#7 .918 & \#6 .991 & \#3 .633 & \#1 .501 & \#6 .505 & \textbf{\#3 .488} \\
& \texttt{deepseek/deepseek-r1-0528} & \#8 .960 & \#1 .940 & \#5 .991 & \#1 .734 & \#2 .500 & \#2 .542 & \textbf{\#4 .482} \\
& \texttt{google/gemini-3-flash-preview} & \#1 .978 & \#2 .936 & \#2 .995 & \#5 .465 & \#5 .489 & \#4 .537 & \textbf{\#5 .478} \\
& \texttt{qwen/qwen3-30b-a3b-thinking-2507} & \#7 .963 & \#5 .921 & \#7 .989 & \#6 .442 & \#6 .478 & \#5 .519 & \textbf{\#6 .467} \\
& \texttt{openai/gpt-5} & \#2 .973 & \#6 .918 & \#1 .995 & \#7 .361 & \#7 .415 & \#7 .465 & \textbf{\#7 .404} \\
& \texttt{qwen/qwen3-235b-a22b} & \#3 .973 & \#8 .807 & \#8 .834 & \#8 .208 & \#8 .286 & \#8 .331 & \textbf{\#8 .314} \\
\midrule
\multirow{8}{*}{MMLU}
& \texttt{openai/gpt-5-mini} & \#2 .940 & \#1 .973 & \#2 .994 & \#1 .911 & \#1 .549 & \#1 .637 & \textbf{\#1 .549} \\
& \texttt{google/gemini-3.1-pro} & \#1 .945 & \#2 .961 & \#1 .999 & \#4 .526 & \#3 .478 & \#2 .584 & \textbf{\#2 .489} \\
& \texttt{qwen/qwen3.6-plus} & \#3 .940 & \#5 .933 & \#3 .991 & \#6 .331 & \#4 .461 & \#3 .582 & \textbf{\#3 .488} \\
& \texttt{openai/gpt-5.4-nano} & \#6 .915 & \#3 .948 & \#4 .980 & \#2 .699 & \#2 .488 & \#4 .551 & \textbf{\#4 .486} \\
& \texttt{x-ai/grok-4.3} & \#4 .940 & \#4 .939 & \#5 .974 & \#3 .649 & \#5 .440 & \#5 .545 & \textbf{\#5 .421} \\
& \texttt{deepseek/deepseek-chat} & \#7 .835 & \#7 .834 & \#6 .927 & \#5 .356 & \#6 .397 & \#6 .451 & \textbf{\#6 .395} \\
& \texttt{anthropic/claude-3.5-haiku} & \#8 .710 & \#8 .726 & \#8 .847 & \#8 .211 & \#7 .309 & \#7 .332 & \textbf{\#7 .282} \\
& \texttt{minimax/minimax-m2.7} & \#5 .930 & \#6 .888 & \#7 .897 & \#7 .317 & \#8 .297 & \#8 .305 & \textbf{\#8 .251} \\
\bottomrule
\end{tabular}
}
\caption{
Detailed leaderboards across all benchmarks.
Each cell reports rank and normalized score under the corresponding evaluation protocol.
}
\label{tab:app-all-leaderboards}
\end{table*}

\end{document}